\documentclass[letterpaper,10pt]{article}

\usepackage{url}
\usepackage[utf8]{inputenc}
\usepackage{lmodern}
\usepackage{graphicx}
\usepackage{amssymb,amsmath}
\usepackage[T1]{fontenc}
\usepackage[hmargin=1in, vmargin=1in]{geometry}
\usepackage{times}
\usepackage{subcaption}
\usepackage{tabularx}
\usepackage{hyperref}

\usepackage{wrapfig}

\usepackage{array}
\usepackage[shortlabels]{enumitem}
\setitemize{itemsep=5pt,topsep=5pt,parsep=0pt,partopsep=5pt,leftmargin=40pt}
\setenumerate{itemsep=5pt,topsep=5pt,parsep=0pt,partopsep=5pt,leftmargin=40pt}
\usepackage{xcolor}

\usepackage{natbib}
\bibpunct{(}{)}{;}{a}{,}{,}

\pdfoutput=1
\hypersetup{
    colorlinks=true,
    linkcolor=black,
    citecolor=black,
    filecolor=black,
    urlcolor=black,
}

\newcommand{\ignore}[1]{}

\title{Latin BERT: \\A Contextual Language Model for Classical Philology\vspace{15pt}}
\author{David Bamman\\
School of Information\\
University of California, Berkeley\\ 
\and
Patrick J. Burns\\
Department of Classics\\
University of Texas at Austin\\
}
\date{}

\begin{document}
\maketitle

\begin{abstract}
We present Latin BERT, a contextual language model for the Latin language, trained on 642.7 million words from a variety of sources spanning the Classical era to the 21st century. In a series of case studies, we illustrate the affordances of this language-specific model both for work in natural language processing for Latin and in using computational methods for traditional scholarship: we show that Latin BERT achieves a new state of the art for part-of-speech tagging on all three Universal Dependency datasets for Latin and can be used for predicting missing text (including critical emendations); we create a new dataset for assessing word sense disambiguation for Latin and demonstrate that Latin BERT outperforms static word embeddings; and we show that it can be used for semantically-informed search by querying \emph{contextual} nearest neighbors. We publicly release trained models to help drive future work in this space.

\end{abstract}

\section{Introduction}

The rise of contextual language models\ \citep{peters-etal-2018-deep,devlin-etal-2019-bert} has transformed the space of natural language processing, raising the state of the art for a variety of tasks---including parsing\ \citep{Mrini2019RethinkingST}, word sense disambiguation\ \citep{huang-etal-2019-glossbert} and coreference resolution\ \citep{joshi-etal-2020-spanbert}---and enabling new ones altogether. While BERT was initially released with a model for English along with a multilingual model trained on aggregated data in 104 languages (mBERT), subsequent work has trained language-specific models for Chinese\ \citep{chinese-bert-wwm}, French\ \citep{martin-etal-2020-camembert,le2019flaubert}, Russian\ \citep{kuratov2019adaptation}, Spanish\ \citep{CaneteCFP2020} and at least 14 other languages, demonstrating substantial improvements for several NLP tasks compared to an mBERT baseline\ \citep{nozza2020what}. In this work, we contribute to this growing space of language-specific models by training a BERT model for the Latin language.

In many ways, research in Latin (and Classics in general) has been at the forefront of computational research for historical languages \citep{berti_digital_2019}. Some of the earliest digitized corpora in the humanities concerned Latin works (Roberto Busa's \emph{Index Thomisticus} of the works of Thomas Aquinas); and one of the flagships of early work in the digital humanities, the Perseus Project\ \citep{crane1996building,smith_perseus_2000}, is structured around a large-scale digital library of primarily Classical texts. At the specific intersection of Classics and NLP, Latin has been the subject of several dependency treebanks \citep{bamman06,haug08,passarotti_improvements_2010,cecchini_new_2020} and other lexico-semantic resources \citep{mambrini_lila_2020,short_latin_2020}, and is the focus of much work on individual components of the NLP pipeline, including lemmatizers, part-of-speech taggers, and morphological analyzers, among others (for overviews, see \citet{mcgillivray_methods_2014} and \citet{burns_building_2019}). This work on corpus creation and annotation as well as the development of NLP tools has enabled literary-critical work on problems relevant to historical-language texts, including uncovering instances of intertextuality in Classical texts\ \citep{coffee2012,moritz-etal-2016-non,coffee_agenda_2018} and stylometric research on genre and authorship \citep{dexter_quantitative_2017,chaudhuri_small_2019,kontges_measuring_2020,storey_like_2020}.

Research in Latin has also made use more specifically of recent advancements in word embeddings, primarily using static word representations such as word2vec\ \citep{mikolov13} to drive work in this space. In the context of NLP, this work includes using lemmatized word embeddings on synonym tasks \citep{sprugnoli_vir_2019} as well as using a variety of embeddings strategies to improve the tasks of lemmatization and POS tagging \citep{sprugnoli_overview_2020} and in particular to improve cross-temporal and cross-generic performance on these tasks \citep{celano_gradient_2020,bacon_data-driven_2020,straka_udpipe_2020}. \citet{bloem_distributional_2020} look at the performance of Latin word embedding models in learning domain-specific sentence representations, specifically sentences taken from Neo-Latin philosophical texts. In literary critical contexts, \citet{bjerva_word_2015,bjerva_rethinking_2016} and \citet{manjavacas_feasibility_2019} have used embeddings to model intertextuality and allusive text reuse. Distributional semantics has received attention elsewhere in Classics, including the work of \citet{rodda_vector_2019}, which uses an Ancient Greek vector space model to explore semantic variation in Homeric formulae. It should also be noted that Latin is often included in large, multilingual NLP studies\ \citep{ammar_massively_2016}; the existence of Vicipaedia, a Latin-language version of Wikipedia, 
has led to the language's inclusion in published multilingual embedding collections, including FastText\ \citep{grave2018learning} and mBERT \citep{devlin-etal-2019-bert}.

In this work, we expand on this existing focus on word representations to build a new BERT-based contextual language model for Latin, trained on 642.7 million tokens from a range of sources, spanning the Classical era through the present. Our work makes the following contributions:

\begin{itemize}
    \item We openly publish a new BERT model for Latin, trained on a dataset of 642.7 million tokens.
    \item We demonstrate new state-of-the-art performance for Latin POS tagging on all three Universal Dependency datasets.
    \item We create a new dataset for assessing word sense disambiugation in Latin, using data from the \emph{Latin Dictionary} of \citet{lewisandshort}.
    \item We illustrate the affordances of Latin BERT for applications in NLP and digital Classics with four case studies, including text infilling and finding contextual nearest neighbors.
\end{itemize}

\noindent
Code and data to support this work can be found at \url{https://github.com/dbamman/latin-bert}.
\section{Corpus}

\begin{wraptable}{h!}{3in}
\begin{centering}
\begin{tabular}{| l | r |} \hline
Source&Tokens \\ \hline \hline
Corpus Thomisticum&14.1M\\ \hline
Internet Archive&561.1M\\ \hline
Latin Library&15.8M\\ \hline
Patrologia Latina&29.3M\\ \hline
Perseus&6.5M\\ \hline
Latin Wikipedia&15.8M\\ \hline \hline
Total&642.7M\\ \hline
\end{tabular}
\caption{\label{corpus} Corpus for Latin BERT.}
\end{centering}
\end{wraptable}
Contextual language models demand large corpora for pre-training: English\ \citep{devlin-etal-2019-bert} and Spanish\ \citep{CaneteCFP2020}, for example, are trained on 3 billion words, while the French CamemBERT model is trained on 32.7 billion\ \citep{martin-etal-2020-camembert}. While Latin is a historical language and is comparatively less resourced than modern languages, there are extant works written in the language covering a time period of over twenty-two centuries---from 200 BCE to the present---resulting in a wealth of textual data \citep{stroh_latein_2007,leonhardt_latin_2013}. In order to capture this variation in usage, we leverage data from several sources: texts from the Perseus Project, which primarily covers the Classical era; the Latin Library, which covers the full chronological scope of the language; the Patrologia Latina, which covers ecclesiastical writers from the 3rd century to the 13th century CE; the Corpus Thomisticum, which covers the (voluminous) writings of Thomas Aquinas; Latin Wikipedia (Vicipaedia), which contains articles on a wide variety of subjects, including contemporary subjects like \emph{Star Wars}, Serena Williams, and Wikipedia itself, written in Latin; and texts from the Internet Archive (IA), which contain a total of 1 billion words spanning works published between roughly 200 BCE and 1922 CE\ \citep{bamman2011b}. The Internet Archive texts are OCR'd scans of books and contain varying OCR quality; in order to use only data with reasonable quality, we retain only those books where at least 40\% of tokens are present in a vocabulary derived from born-digital texts. Table \ref{corpus} presents a summary of the corpus and its individual components; since the texts from the Internet Archive are noisier than the other subcorpora, we uniformly upsample all non-IA texts to train on a balance of approximately 50\% IA texts and 50\% non-IA texts.

We tokenize all texts using the same Latin-specific tokenization methods in the Classical Language Toolkit \citep{johnson_classical_2020}, both for delimiting sentences and tokenizing words; the CLTK word tokenizer segments enclitics from their adjoining word so that \emph{arma virumque cano} (``I sing of arms and the man'') is tokenized into [\emph{arma}], [\emph{virum}], [\emph{-que}], [\emph{cano}]. Since BERT operates on subtokens rather than word tokens, we learn a Latin-specific WordPiece tokenizer using the {tensor2tensor} library from this training data, with a resulting vocabulary size of 32,895 subword types. This method tokenizes \emph{audentes fortuna iuvat} (``fortune favors the daring'') into the sequence [\emph{audent}, \emph{es}], [\emph{fortuna}], [\emph{iuvat}] in order for representations to be learned for subword units rather than for the much larger space of all possible inflectional variants of a word, which substantially reduces the representation space for highly inflected languages like Latin. In the experiments that follow, we generate a BERT representation for a token comprised of multiple WordPiece subtokens (such as \emph{audentes} above) by averaging its component subtoken representations.

\section{Training}

Our Latin BERT model contains 12 layers and a hidden dimensionality of 768; we pre-train it with whole word masking using tensorflow on a TPU for one million steps. Training took approximately 5 days on a TPU v2, and cost \$540 on Google Cloud (at \$4.50 per TPU v2 hour). We set the maximum sequence length to 256 WordPiece tokens and trained with a batch size of 256. Details on other hyperparameter settings can be found in the Appendix. We convert the resulting tensorflow checkpoint into a BERT model that can be used by the HuggingFace library; the trained model and code to support it can be found at \url{https://github.com/dbamman/latin-bert}.

\section{Analyses}

A Latin-specific contextual language model offers a range of affordances for work both in Classics NLP (improving the state of the art for various NLP tasks for the domain of Latin) and in using computational methods to inform traditional scholarly analysis. We present four case studies to illustrate these possibilities: improving POS tagging for Latin (yielding a new state-of-the-art for the UD datasets); predicting editorial reconstructions of texts; disambiguating Latin word senses in context; and enabling \emph{contextual nearest neighbor} queries---that is, finding specific passages containing words that are used in similar contexts to a given query.

\subsection{POS tagging}

BERT has been shown to learn representations that encode many aspects of the traditional NLP pipeline, including POS tagging, parsing, and coreference resolution\ \citep{tenney-etal-2019-bert,hewitt2019structural}. To explore the degree to which Latin BERT can be useful for the individual stages in NLP, we focus on POS tagging. POS tagging is an important component in much work in Latin NLP, providing the scaffolding for dependency parsing and detecting text reuse\ \citep{moritz-etal-2016-non}, and providing a focus for the 2020 EvaLatin shared task\ \citep{sprugnoli_overview_2020}.

\begin{figure}[ht]
\begin{centering}
\includegraphics[scale=.75]{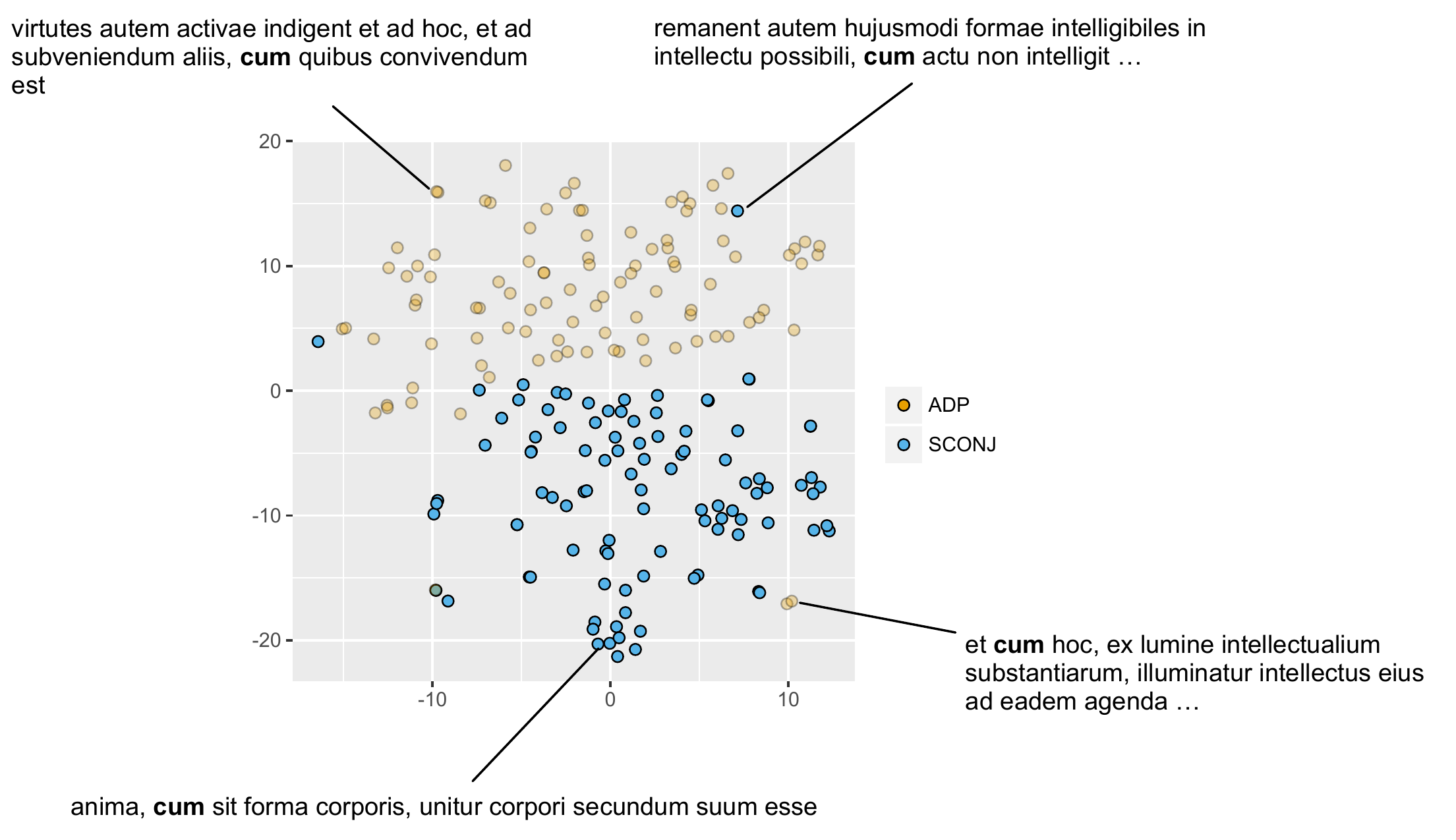}
\caption{\label{cum} Part of speech distinctions for \emph{cum} as preposition (ADP) vs. subordinating conjunction (SCONJ), along with examples of each class.}
\end{centering}
\end{figure}

To understand how contextual representations in BERT capture morphosyntactic information, we can examine a case study of the ambiguity present in the frequent word form \emph{cum}, a homograph with two distinct meanings: it is used both as a preposition appearing with a nominal in the ablative case (meaning \emph{with}) and as a subordinating conjunction (meaning \emph{when}, \emph{because}, \emph{although}, etc.). To illustrate the degree to which raw BERT representations naturally encode this distinction, we sample 100 sentences containing \emph{cum} as a preposition (ADP) and 100 sentences with it as a subordinating conjunction (SCONJ) from the Index Thomisticus Treebank\ \citep{passarotti_improvements_2010}, run all sentences through Latin BERT, and generate a representation of each instance of \emph{cum} as the final layer of BERT at that token position (each \emph{cum} is therefore represented as a 768-dimensional vector). In order to visualize the results in two dimensions, we then carry out dimensionality reduction using t-SNE\ \citep{maaten2008visualizing}. Figure \ref{cum} illustrates the result: the representations of \emph{cum} as a preposition and as a subordinating conjunction are nearly perfectly separable, indicating that the distinction between the use of \emph{cum} corresponding to these parts of speech is inherent in its contextual representation within BERT without any further training necessary to tailor it to POS tagging.\footnote{Two of the instances of \emph{cum}/ADP that cluster with SCONJ are in the collocation \emph{cum hoc}; this collocation appears 17 times in the ITT data and in 14 of these instances \emph{cum} is labelled SCONJ. The third instance of \emph{cum}/ADP that clusters with SCONJ (\emph{Subdit autem, qui \textbf{cum} in forma Dei esset, non rapinam arbitratus est esse se aequalem Deo.}) is mislabelled in the data; accordingly, it is in the correct cluster. As far as the instances of \emph{cum}/SCONJ that cluster with ADP, one is followed by an ablative noun (\emph{cum actu}), which perhaps affects its representation; the only other instance of \emph{cum actu} in the ITT data has \emph{cum} labelled as ADP. The second instance (\emph{Ergo \textbf{cum} aliae formae sint simplices, multo fortius anima.}) provides no clue as to its misclassification.}

While this case study provides a measure of face validity on the ability of Latin BERT to meaningfully distinguish major sense distinctions for a frequent word---without explicitly being trained to do so---we can also test its capacity to be used for the specific task of POS tagging. To do so, we draw on three different dependency treebanks annotated with morphosyntactic information: the Perseus Latin Treebank\ \citep[Perseus]{bamman06}, containing works from the Classical period (18,184 training tokens); the Index Thomisticus Treebank\ \citep[IITB]{passarotti_improvements_2010}, containing works by Thomas Aquinas (293,305 training tokens); and the PROIEL treebank\ \citep[PROIEL]{haug08}, containing both Classical and Medieval works (172,133 training tokens). We build a POS tagger by adding a linear transformation and softmax operation on top of the pre-trained Latin BERT model, and allowing all of the parameters within the model to be fine-tuned during training. For ITTB and PROIEL, early stopping was assessed on performance on development data (lack of improvement after 10 iterations), while the Perseus model (which has no development split due to its size) was trained for a fixed number of 5 epochs.

We compare performance with several alternatives.
First, to contextualize performance with static word representations, we also train 200-dimensional static word2vec embeddings\ \citep{mikolov13} for Latin using the same training data as Latin BERT, and use these as trainable word representations in a bidirectional LSTM (\emph{static embeddings} below); to compare performance with a similar model that uses the multilingual mBERT---trained simultaneously on different versions of Wikipedia in many different languages---we report test accuracy from\ \citet{straka2019evaluating}. And to provide context from several other static systems at the the 2018 Shared Task on universal dependency parsing (which includes a subtask on POS tagging on these datasets), we report test accuracies from \citet{smith-etal-2018-82}, \citet{straka-2018-udpipe} and \citet{boros-etal-2018-nlp}.

As can be seen, Latin BERT generates a new state of art for POS tagging on these datasets, with the most dramatic improvement coming in its performance on the small Perseus dataset (an improvement of 4.6 absolute points).

\begin{table}[h]
\begin{centering}
\begin{tabular}{| l | c | c | c |} \hline
Method&Perseus&PROIEL&ITTB \\ \hline \hline
Latin BERT&\textbf{94.3}&\textbf{98.2}&\textbf{98.8} \\ \hline
\citet{straka2019evaluating}&90.0&97.2&98.4 \\ \hline
\citet{smith-etal-2018-82}&88.7&96.2&98.3 \\ \hline
\citet{straka-2018-udpipe}&87.6&96.8&98.3 \\ \hline
Static embeddings&87.8&95.2&97.7 \\ \hline
\citet{boros-etal-2018-nlp}&85.7&94.6&97.7 \\ \hline
\end{tabular}
\caption{\label{postagging} POS tagging results.}
\end{centering}
\end{table}

\subsection{Text infilling}

One of BERT's primary training objectives is \emph{masked language modeling}---randomly selecting a word from an input sentence and predicting it from representations of the surrounding words. This inductive bias makes it a natural model for the task of text infilling\ \citep{zhu2019text,assael-etal-2019-restoring,donahue-etal-2020-enabling}, in which a model is tasked with predicting a word (or sequence of words) that has been elided in some context.

Previous work has largely used synthetic evaluations for this task---both for predicting missing words in English and  Ancient Greek\ \citep{assael-etal-2019-restoring}---by randomly masking a word in a complete sentence and attempting to predict it; to more closely align this task with the scholarly practice of textual criticism, in which an editor reconstructs a text that has been corrupted or is otherwise illegible, we create an evaluation set by exploiting orthographic markers of emendation---specifically, the angle brackets (< and >) typically used to mark ``words ... added to the transmitted text by conjecture or from a parallel source'' \citep[80]{west_textual_1973}. In the following sentence, for example, an editor notes that the word \emph{ter} is a conjecture by surrounding it in angle brackets: 

\begin{quote}
\begin{center}
    populus romanus	<ter> cum carthaginiensibus dimicavit.\footnote{``The Roman people fought against the Carthaginians <three times>,'' Ampelius, \emph{Liber Memorialis} 46.}
    \end{center}
\end{quote}

We build a dataset of textual emendations by searching all texts in the Latin Library for single words at least two characters long surrounded by brackets in sentences ranging between 10 and 100 words in length. In order to ensure that the emendation does not appear in BERT's training data, we first removed all sentences with single words in angle brackets from the training data prior to training BERT, and removed any evaluation sentence whose 5-gram centered on the conjecture appeared in the training data (since emended text may appear in other versions of the text without explicit markers of the emendation). In the example above, if the 5-gram \emph{populus romanus ter cum carthaginiensibus} appeared in the training data, we would exclude this example from evaluation.

The resulting evaluation dataset contains 2,205 textual emendations. We find that Latin BERT is able to reconstruct the human-judged emendation 33.1\% of the time; in 62.2\% of cases, the human emendation is in the top 10 predictions ranked by their probability, and in 74.0\% of cases, it is within the top 50. Table \ref{emendations} illustrates several examples of successes and failures for this task.

\begin{table}[h]
\begin{centering}
\begin{tabular}{| p{6cm} | c | c | p{6cm} |} \hline
Left context&Prediction&Emendation&Right context \\ \hline \hline praetorius qui bello civili partes pompei secutus mori maluit quam superstes esse rei&publicae&publicae&servienti. \\ \hline
hanno et mago qui&secundo&primo&punico bello cornelium consulem aput liparas ceperunt. \\ \hline
tiberis infestiore quam priore&anno&anno&impetu inlatus urbi duos pontes, aedificia multa maxime circa flumentanam portam euertit. \\ \hline
brachium enim tuum non&dominus&domini&dixisset, si non dominum patrem et dominum filium intellegi vellet. \\ \hline
postquam dederat universitati parem dignamque faciem, spiritum desuper, quo pariter&omnes&omnia &animarentur, inmisit. \\ \hline
\end{tabular}
\caption{\label{emendations} Examples of infilling textual emendations.}
\end{centering}
\end{table}

\begin{wraptable}{h!}{7cm}
\begin{centering}
\begin{tabular}{| l | r |} \hline
Word&Probability \\ \hline \hline
secundo&0.451\\ \hline
primo&0.385\\ \hline
tertio&0.093\\ \hline
altero&0.018\\ \hline
primi&0.012\\ \hline
priore&0.012\\ \hline
quarto&0.005\\ \hline
secundi&0.004\\ \hline
primum&0.002\\ \hline
superiore&0.002\\ \hline
\end{tabular}
\caption{\label{ranked} Candidate words filling ``hanno et mago qui \textunderscore\textunderscore\textunderscore\; punico bello cornelium consulem aput liparas ceperunt,'' ranked by their probability.}
\end{centering}
\end{wraptable} In addition to providing a single best prediction for a missing word, language models such as BERT produce a probability distribution over the entire vocabulary, and we can use those probabilities to generate a ranked list of candidates. Table \ref{ranked} below presents one such list of ranked candidates to fill the missing word in ``hanno et mago qui \textunderscore\textunderscore\textunderscore\; punico bello cornelium consulem aput liparas ceperunt'' (``Hanno and Mago, who captured the consul Cornelius at Lipari in the [blank] Punic War''); while Latin BERT predicts \emph{secundo} as its best guess, the human emendation of \emph{primo} also ranks highly. This is especially interesting as it reflects to some degree the editor’s decision-making process. According to Eduard Wölfflin's 1854 Teubner edition, Karl Halm added \emph{primo} to a section of Ampelius's \emph{Liber Memorialis} on Carthaginian generals and kings. His addition to the text clearly places the battle at Lipari during which Cornelius (i.e. Gnaeus Cornelius Scipio Asina) was captured as having taken place during the First Punic War. Yet the collocation of Hanno et Mago (at least in the extant texts available for use as training data; e.g. Livy 23.41, 28.1; Val. Max. 7.2 ext 16; Sil. \emph{Pun.} 16.674) is more closely associated with the Second Punic War and it is perhaps the case that Ampelius has misreported the subject of this sentence. So with respect to contextual semantics, the left context connotes ``Second Punic War'' and the right context connotes ``First Punic War.’’ Accordingly, Halm must draw on external, historical information (i.e. the dating of Lipari) to establish what semantic context alone cannot. Latin BERT, on the other hand, unable to draw on external information, makes a reasonable suggestion based on what is reported in Ampelius’s text and how his words relate to other texts.

\subsection{Word Sense Disambiguation}

Many word forms in Latin have multiple senses, both at the level of homographs (words derived from distinct lemmas, such as \emph{est}, which can be derived both from the verb \emph{edo} (``to eat'') and the far more common \emph{sum} (``to be'') and within a single dictionary entry as well---the word \emph{in}, for example, is a preposition that denotes both location \emph{within} (where it appears with nouns in the ablative case) and movement \emph{towards} (where it appears with nouns in the accusative case). Most ambiguous words exhibit sense variation without direct impact on the morphosyntactic structure of the surrounding sentence (where, for example, the word \emph{vir} can refer to ``a man'' generally and also the more specific ``husband'').

Within NLP, word sense disambiguation provides a mechanism for assigning a word in context to a choice of sense from a pre-defined sense inventory. Broad-coverage WSD system are typically evaluated on annotated datasets such as those in English from the Senseval and SemEval competitions\ \citep{raganato-etal-2017-word}, where human annotators select the appropriate sense for a word given its context of use within a sentence. While these datasets exist for languages like English, they have yet to be created for Latin.

\begin{wrapfigure}{h!}{3.5in}
\begin{centering}
\includegraphics[scale=.75]{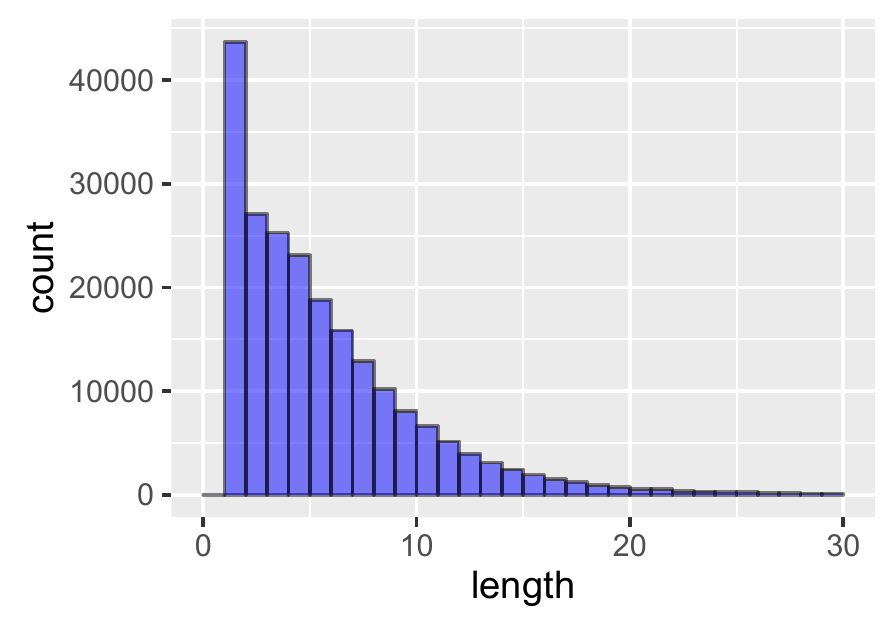}
\caption{\label{citlen} Distribution of citation lengths in Lewis and Short.}
\end{centering}
\vspace{-5pt}
\end{wrapfigure}
\begin{figure}[h]
\begin{centering}
\includegraphics[scale=.5]{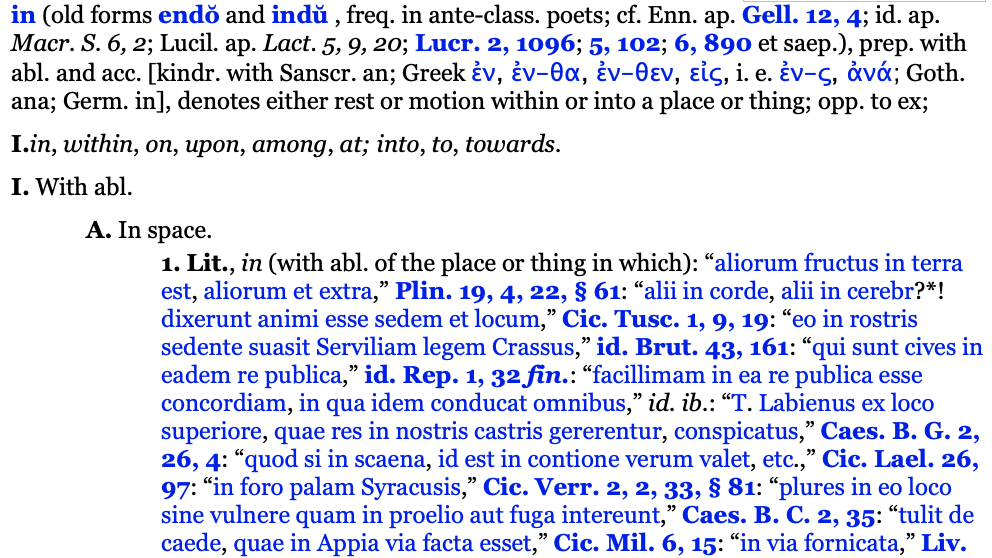}
\caption{\label{lewisin} Selection from the Lewis and Short entry for \emph{in} (Perseus Digital Library).}
\end{centering}
\end{figure}

In order to explore the affordances of BERT for word sense disambiguation, we create a new WSD dataset for Latin by mining sense examples from the \emph{Latin Dictionary} of \citet{lewisandshort}, which provides both a sense inventory for Latin words and a set of attestations of those senses primarily in Classical authors. Figure \ref{lewisin} provides one such example of the dictionary entry for the preposition \emph{in} within this dictionary, illustrating its first sense along with attestations of its use.
As figure \ref{citlen} illustrates, the majority of example sentences are fragmentary in nature, with 55\% of them having a length fewer than 5 words. We build a dataset from this source by selecting dictionary headwords that have at least two distinct major senses (denoted by ``I.'' and ``II.'' typographically) that are supported by at least 10 example sentences, where each sentence is longer than 5 words to provide enough context for disambiguation. We transform the problem into a balanced binary classification for each headword by only selecting the first two major senses, and balancing the number of examples for each sense to be equal. This results in a final dataset comprised of 8,354 examples for a total of 201 dictionary headwords. We divide this data into training (80\%), development (10\%) and test splits (10\%) for each individual headword, preserving balance across all splits.  

We evaluate Latin BERT on this dataset by fine-tuning a separate model for each dictionary headword; the number of training instances per headword ranges from 16 (8 per sense) to 192 (96 per sense); 59\% of headwords have 24 or fewer training examples.
For comparison, we also present the results of a 200-dimensional bidirectional LSTM with static word embeddings as input. For both models, we determine the number of epochs to train on based on performance on development data, and report accuracy on held-out test data.

Table \ref{wsd} presents these results: while random choice would result in 50\% accuracy on this balanced dataset, static embeddings achieve an overall accuracy of 67.3\%. Even when presented with only a few training examples, however, Latin BERT is able to learn meaningful sense distinctions, yielding a 75.4\% accuracy (an absolute improvement of 8.1 points over a non-contextual model).

\begin{table}[h!]
\begin{centering}
\begin{tabular}{| l | r |} \hline
Method&Accuracy \\ \hline \hline
Latin BERT&75.4\% \\ \hline
Static embeddings&67.3\% \\ \hline
Random&50.0\% \\ \hline
\end{tabular}
\caption{\label{wsd} WSD results.}
\end{centering}
\end{table}

While the absolute performance of these models is naturally not as strong as those for POS tagging, this reflects the difficulty of word sense disambiguation as a task (where, for comparison, it is only recently that BERT-based models in English have been able to demonstrate significant improvements over a most frequent sense baseline\ \citep{huang-etal-2019-glossbert}). This experimental design also points the way for similar work in other Classical languages like Ancient Greek, which have similar lexica (such as the \emph{LSJ Greek-English Lexicon}\ \citep{lsj}) that contain a variety of examples for each dictionary sense. And while we focus in this evaluation on citation examples longer than five words, this work could be significantly expanded to include far more training and evaluation examples by retrieving full-text examples from the fragmentary citations.

\subsection{Contextual Nearest Neighbors}

One final case study that we can examine is the use of contextual word embeddings to find similar passages to a query. While static word embeddings like word2vec\ \citep{mikolov13} and GloVe\ \citep{pennington2014glove} allow for the comparison of nearest neighbors, similarity in those models is only scoped over word \emph{types}, and not over specific instances of those words as tokens in context. Finding similar tokens in context would enable a range of applications in digital Classics, including discovering instances of intertextuality (where, for example, the contextual representations for tokens in Ovid's \emph{arma gravi numero violentaque bella parabam / edere} (``I was getting ready to publish something in a serious meter about arms and violent wars'')  may bear some similarity to Vergil's \emph{arma virumque cano}), surfacing examples for pedagogy (where an instructor may want to find examples of ablative absolute constructions in extant Latin texts by finding passages that are similar to \emph{his verbis dictis} (``with these words having been said'')), or suggesting to the textual critic additional apposite ``parallel'' passages when reconstructing texts. We illustrate the potential here by generating Latin BERT representations for 16 million tokens of primarily Classical Latin texts, and finding the nearest neighbor for a query token representation.
While some work using BERT has calculated sentence-level similarity using the representations for the starting [CLS] token\ \citep{qiao2019understanding}, we find that comparing individual tokens within sentences yields much greater precision, since similar subphrases may be embedded within longer sentences that are otherwise dissimilar.

Tables \ref{in} and \ref{audentes} present the contextual nearest neighbor results for two queries: tokens most similar to \textbf{in} in \emph{gallia est omnis divisa {in} partes tres} (``The entirety of Gaul is divided into three parts''), and tokens most similar to \textbf{audentes} in \emph{audentes fortuna iuvat}. The most similar tokens to the first query example within the rest of the corpus not only capture the specific morphological constraints of this sense of \emph{in} appearing with a noun in the accusative case (denoting \emph{into} rather than \emph{within}) but also broadly capture the more specific subsense of division into smaller components---including division into ``parts'' (\emph{partis}, \emph{partes}), ``districts'' (\emph{pagos}), ``provinces'' (\emph{provincias}), units of measurement (e.g. \emph{uncias}) and ``kinds'' (\emph{genera}).

\begin{table}[h!]
\begin{centering}

\begin{tabular}{| c | p{11cm}  | l |} \hline
Cosine&Text&Citation \\ \hline
0.835&ager romanus primum divisus \textbf{in} partis tris, a quo tribus appellata titiensium ...&Varro, \emph{Ling}. \\ \hline
0.834&in ea regna duodeviginti dividuntur \textbf{in} duas partes.&Solin. \\ \hline
0.833&gallia est omnis divisa \textbf{in} partes tres, quarum unam incolunt belgae, aliam ...&Caes., \emph{BGall}.\\ \hline
0.824&is pagus appellabatur tigurinus; nam omnis civitas helvetia \textbf{in} quattuor pagos divisa est.&Caes., \emph{BGall}. \\ \hline
0.820&ea pars, quam africam appellavimus, dividitur \textbf{in} duas provincias, veterem et novam, discretas fossa ...&Plin. \emph{HN} \\ \hline
0.817&eam distinxit \textbf{in} partes quatuor.&Erasmus, \emph{Ep}. \\ \hline
0.812&hereditas plerumque dividitur \textbf{in} duodecim uncias, quae assis appellatione continentur.&Justinian, \emph{Inst}. \\ \hline

\end{tabular}
\caption{\label{in} Most similar tokens to ``in'' in \emph{gallia est omnis divisa in partes tres}.}
\end{centering}
\end{table}

The most similar tokens to \emph{audentes} in \emph{\textbf{audentes} fortuna iuvat} include other versions of the same phrase with lexical variation (\emph{\textbf{audentis} fortuna iuvat}, \emph{\textbf{audaces} fortuna iuvat}) and instances of intertextuality (\emph{\textbf{audentes} forsque deusque iuvat}). This example in particular not only illustrates the ability of BERT to capture meaningful similarities between different instances of the same word (such as \emph{in} in the first example), but also between words that exhibit semantic similarity in spite of surface differences (\emph{audentes}, \emph{audaces} and \emph{audentis}).

\begin{table}[h!]
\begin{centering}
\begin{tabular}{| c | p{11cm}  | l |} \hline
Cosine&Text&Citation\\ \hline
0.926&\textbf{audentes} forsque deusque iuvat.&Ov., \emph{Fast}.\\ \hline
0.864&\textbf{audentis} fortuna iuvat, piger ipse sibi opstat.&Sen., \emph{Ep}.\\ \hline
0.846&\textbf{audentes} in tela ruunt.&Vida, \emph{Scacchia Ludus} \\ \hline
0.840&\textbf{audentes} facit amissae spes lapsa salutis, succurruntque duci&Vida, \emph{Scacchia Ludus}\\ \hline
0.837&... \textbf{audentis} fortuna iuuat.' haec ait, et cum se uersat ...&Verg., \emph{Aen}. \\ \hline
0.815&\textbf{cedentes} urget totas largitus habenas liuius acer equo et turmis ...&Sil. \\ \hline
0.809&sors aequa \textbf{merentes} respicit.&Stat., \emph{Theb}. \\ \hline
0.801&nec jam \textbf{pugnantes} pro caesare didius audax hortari poterat, nec in ...&May, \emph{Supp. Pharsaliae} \\ \hline
0.800&et alibi idem dixit,'' \textbf{audaces} fortuna iuvat, piger sibiipsi obstat.''&Albertanus of Brescia \\ \hline
0.796&quae saeua repente \textbf{uictores} agitat leto iouis ira sinistri?&Stat., \emph{Theb}. \\ \hline
\end{tabular}
\caption{\label{audentes} Most similar tokens to ``audentes'' in \emph{audentes fortuna iuvat}.}
\end{centering}
\end{table}

\section{Conclusion}

In this work, we present the first contextual language model for a historical language, training a BERT-based model for Latin on 642.7 million tokens originally written over a span of 22 centuries (from 200 BCE to today). This large-scale language model has proven valuable for a range of applications, including specific subtasks in the NLP pipeline (including POS tagging and word sense disambiguation) and has the potential to be instrumental in driving forward traditional scholarship by providing estimates of word probabilities to aid in the work of textual emendation and by operationalizing a new form of semantic similarity in contextual nearest neighbors.

While this work presents Latin BERT and illustrates its usefulness with several case studies, there are a variety of directions this work can lead. One productive area of research within BERT-based NLP has been in designing probing experiments to tease apart what the different layers of BERT have learned about linguistic structure; while the BERT representations used as features for several aspects of this work come from the single final layer, this line of research may shed light on which layers are more appropriate for specific tasks (including the best representations for finding comparable passages). While we focus on the tasks of POS tagging and word sense disambiguation in NLP, contextual language models have shown dramatic performance improvements for a range of components in the NLP pipeline, including syntactic parsing and named entity recognition. Given the availability of labeled datasets for these tasks, this is a direction worth pursuit. And finally, we only begin to address the ways in which specifically \emph{contextual} models of language can inform traditional scholarship in Classics; while textual criticism and intertextuality represent two major areas of research where such models can be useful, there are many other potentially fruitful areas where better representation of the contextual meaning of words can be helpful, including Classical lexicography and literary critical applications other than intertextuality detection. We leave it to future work to explore these new directions.

\section*{Acknowledgments}

The research reported in this article was supported by a grant from the National Endowment for the Humanities (HAA256044-17), along with resources provided by the Google Cloud Platform. The authors would also like to acknowledge the Quantitative Criticism Lab and the Institute for the Study of the Ancient World Library for their support.

\bibliographystyle{acl_natbib}
\bibliography{latin_bert}

\clearpage
\section*{Appendix}

\begin{table}[h!]
\begin{centering}
\begin{tabular}{| l | r |} \hline
attention probs dropout prob&0.1\\ \hline
directionality&bidi\\ \hline
hidden act&gelu\\ \hline
hidden dropout prob&0.1\\ \hline
hidden size&768\\ \hline
initializer range&0.02\\ \hline
intermediate size&3072\\ \hline
max position embeddings&512\\ \hline
num attention heads&12\\ \hline
num hidden layers&12\\ \hline
pooler fc size&768\\ \hline
pooler num attention heads&12\\ \hline
pooler num fc layers&3\\ \hline
pooler size per head&128\\ \hline
pooler type&first token transform\\ \hline
type vocab size&2\\ \hline
vocab size&32900\\ \hline
train batch size&256 \\ \hline
max seq length&256 \\ \hline
learning rate&1e-4 \\ \hline
masked lm prob&0.15 \\ \hline
do whole word mask&True \\ \hline
\end{tabular}
\caption{\label{hyperparameters} Latin BERT training hyperparameters.}
\end{centering}
\end{table}

\end{document}